\pdfoutput=1

\documentclass[11pt]{article}

\usepackage[preprint]{acl}
\usepackage{etoolbox}
\newtoggle{isReview}
\togglefalse{isReview}

\usepackage{times}
\usepackage{latexsym}

\usepackage[T1]{fontenc}

\usepackage[utf8]{inputenc}

\usepackage{microtype}

\usepackage{inconsolata}

\usepackage{graphicx}

\usepackage{graphics}
\usepackage{graphicx}
\usepackage{subcaption}
\usepackage{caption}
\usepackage{amssymb}
\usepackage{algorithm}
\usepackage{algpseudocode}
\usepackage{wrapfig}
\usepackage{longtable}
\usepackage{fancyvrb}
\usepackage{empheq}
\usepackage{xspace}
\usepackage{arydshln}
\usepackage{booktabs}
\usepackage{xcolor}

\newcommand{\ours}{RePO}

\title{RePO: Replay-Enhanced Policy Optimization}

\author{
 \textbf{Siheng Li$^{\heartsuit \spadesuit *}$}
 \quad
 \textbf{Zhanhui Zhou$^{\spadesuit}$}\thanks{\ Equal contribution. Work done while Siheng Li and Zhanhui Zhou were at Shanghai AI Lab. Author contributions are listed at the end of the paper.}
 \quad
 \textbf{Wai Lam$^{\heartsuit}$}
 \quad
 \textbf{Chao Yang$^{\spadesuit}$}
 \quad
 \textbf{Chaochao Lu$^{\spadesuit}$}
 \\ 
 $^{\heartsuit}$The Chinese University of Hong Kong
 \\
 $^{\spadesuit}$Shanghai Artificial Intelligence Laboratory
 \\
 {
   \textbf{Correspondence:} \href{mailto:sihengli24@gmail.com}{\texttt{sihengli24@gmail.com}} \quad
   \href{mailto:yangchao@pjlab.org.cn}{\texttt{yangchao@pjlab.org.cn}}
 }
}

\begin{document}
\maketitle

\begin{abstract}
Reinforcement learning (RL) is vital for optimizing large language models (LLMs). Recent Group Relative Policy Optimization (GRPO) estimates advantages using multiple on-policy outputs per prompt, leading to high computational costs and low data efficiency. To address this, we introduce Replay-Enhanced Policy Optimization (RePO), which leverages diverse replay strategies to retrieve off-policy samples from a replay buffer, allowing policy optimization based on a broader and more diverse set of samples for each prompt. Experiments on five LLMs across seven mathematical reasoning benchmarks demonstrate that RePO achieves absolute average performance gains of $18.4$ and $4.1$ points for Qwen2.5-Math-1.5B and Qwen3-1.7B, respectively, compared to GRPO. Further analysis indicates that RePO increases computational cost by $15\%$ while raising the number of effective optimization steps by $48\%$ for Qwen3-1.7B, with both on-policy and off-policy sample numbers set to $8$.
The repository can be accessed at \url{https://github.com/SihengLi99/RePO}.
\end{abstract}
\section{Introduction}
Large language models (LLMs) have made significant strides in aligning with human values \citep{bai2022training, ouyang2022training}, complex reasoning \citep{guo2025deepseek, jaech2024openai}, and autonomous agents \citep{wang2024survey}. A key technique driving these advancements is reinforcement learning (RL), which reinforces behaviors associated with higher rewards while reducing those linked to lower rewards.

Recent RL approaches for LLMs have focused on Group Relative Policy Optimization (GRPO) \citep{shao2024deepseekmath, liu2025understanding, yu2025dapo, lin2025cppo}, which estimates advantages by sampling multiple on-policy outputs per prompt and normalizing their rewards. Although GRPO has shown promising results, it is inherently on-policy and requires multiple on-policy samples per prompt, resulting in substantial computational overhead. Additionally, relying solely on on-policy samples can be limiting; for example, when all samples receive the same rewards, the estimated advantages collapse to zero, thereby providing no meaningful gradient signal for optimization.

\begin{figure*}[htb]
  \centering
  \includegraphics[width=1.0\textwidth]{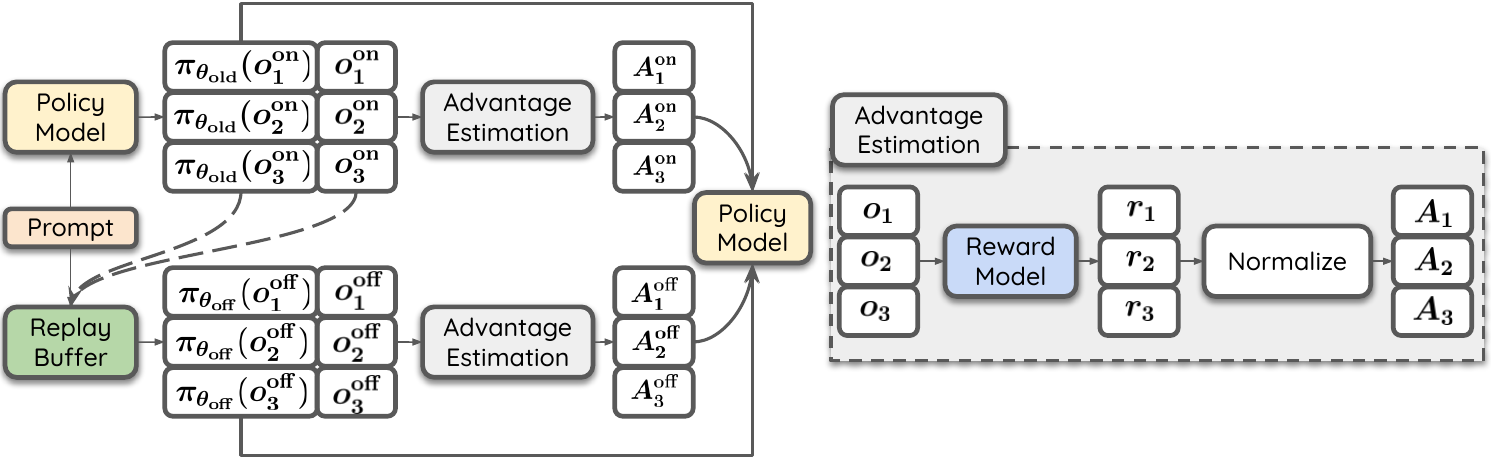}
  \caption{Demonstration of \ours. The policy is updated using both on-policy samples and off-policy samples retrieved from a replay buffer. Advantage estimation is performed separately for on-policy and off-policy updates.}
  \label{fig:method}
\end{figure*}

To address these limitations, we propose Replay-Enhanced Policy Optimization (\ours), an extension of GRPO that integrates both on-policy and off-policy updates. Specifically, \ours\ employs diverse replay strategies to retrieve suitable samples from a replay buffer that stores previously sampled outputs for each prompts. Its objective combines on-policy and off-policy terms, enabling the use of a broader set of outputs per prompt,  which improves data efficiency and reduces overfitting. Additionally, diverse replay strategies offer flexibility in optimization. For instance, high-reward samples can be prioritized to reinforce desirable behaviors, while samples more closely aligned with the current policy can be selected to reduce discrepancies between the behavior and current policies.

We evaluate the effectiveness of \ours\ through experiments on five open-source LLMs. The results indicate that \ours\ outperforms GRPO, achieving absolute average accuracy gains of $18.4$, $2.0$, and $4.1$ points for Qwen2.5-Math-1.5B, Qwen2.5-Math-7B, and Qwen3-1.7B, respectively, on mathematical reasoning tasks. Additionally, the performance on general reasoning benchmarks further underscores the generalization capability of RePO. Analytical studies show that for Qwen3-1.7B, with both on-policy and off-policy sample numbers set to $8$, \ours\ relatively increases computational cost by $15\%$ while raising the number of effective optimization steps by $48\%$. We expect that \ours\ will improve RL optimization for LLMs, contributing to the continual advancement of LLM capabilities.
\section{Preliminary}
\subsection{Reinforcement Learning}
Reinforcement learning (RL) has proven effective in optimizing LLMs. Proximal Policy Optimization (PPO) \citep{schulman2017proximal} is one of the most widely adopted methods for optimizing LLMs. The PPO objective is defined as follows:
\begin{align*}
&\mathcal{J}_{\mathrm{PPO}}(\theta) = \mathbb{E}_{q\sim P(Q),\,o\sim \pi_{\theta_{\mathrm{old}}}(O\mid q)} \\
&  \frac{1}{\lvert o\rvert}\sum_{t=1}^{\lvert o\rvert}
   \begin{aligned}[t]
     \min\bigl[\,r_t\,A_t,\mathrm{clip}\bigl(r_t,\,1-\epsilon,\,1+\epsilon\bigr)\,A_t\bigr].
   \end{aligned}
\end{align*}
Here, $q$ is the prompt, and $o=(o_1,\dots,o_T)$ represents the output token sequence sampled from the behavior policy $\pi_{\theta_{\mathrm{old}}}$, and $$r_t=\frac{\pi_{\theta}(o_t\mid q,o_{<t})}{\pi_{\theta_{\mathrm{old}}}(o_t\mid q,o_{<t})}$$ is the importance‐sampling ratio between the current policy $\pi_{\theta}$ and the behavior policy $\pi_{\theta_{\mathrm{old}}}$. The clipping hyperparameter $\epsilon > 0$ constrains the extent of policy updates, preventing large deviations from the behavior policy. The estimated advantage for the $t$-th token, $A_t$, directs the PPO objective to favor actions with higher advantage estimates while diminishing those with lower advantages. To estimate $A_t$, Generalized Advantage Estimation (GAE) \citep{schulman2015high} is employed, which relies on a large value model and incurs substantial memory and computational overhead.

\subsection{Group Relative Policy Optimization}
\label{sec:grpo}
Group Relative Policy Optimization (GRPO) \citep{shao2024deepseekmath} addresses this issue by sampling multiple outputs $\{o_1, \ldots, o_G\}$ for each prompt and leveraging their rewards $\mathcal{G} = \{R(o_1), \ldots, R(o_G)\}$, where $R(o_i)$ is the reward associated with output $o_i$, to estimate the advantage for each output:
\begin{align*}
    A_{i,t} &= \frac{R(o_i) - \mathrm{mean}(\mathcal{G})}{\mathrm{std}(\mathcal{G})}.
\end{align*}
The objective function is as follows:
\begin{align*}
&\mathcal{J}_{\mathrm{GRPO}}(\theta) = \mathbb{E}_{\,q\sim P(Q),\,\{o_i\}_{i=1}^G \sim \pi_{\theta_{\mathrm{old}}}(O\mid q)} \\
& \frac{1}{G}\sum_{i=1}^G \frac{1}{\lvert o_i\rvert}
   \sum_{t=1}^{\lvert o_i\rvert}
    \Bigl\{
     \min\bigl[
         r_{i,t}\, A_{i,t},\, \\
& \, \mathrm{clip}\bigl(r_{i,t},1-\epsilon,1+\epsilon\bigr)\, A_{i,t}\bigr] - \beta\,D_{\mathrm{KL}}\bigl[\pi_{\theta}\,\|\,\pi_{\mathrm{ref}}\bigr] \Bigl\}
\end{align*}
Here, $\beta$ regulates the KL divergence between the current policy and reference policy. A notable \textbf{limitation} of GRPO is its reliance on multiple on-policy samples for each prompt, resulting in high computational cost. Additionally, if all samples receive the same reward, the advantage $A_{i,t}$ becomes zero, diminishing the optimization signal. This issue is particularly evident in tasks that are overly simple or excessively difficult or when the current policy produces less diverse outputs during training, a common drawback observed in GRPO \citep{yu2025dapo, yue2025does, cui2025entropy}.

\section{Replay-Enhanced Policy Optimization}
\label{sec:repo}
This paper proposes Replay-Enhanced Policy Optimization (RePO), a method that mitigates the dependence on multiple on-policy samples by leveraging previously sampled outputs for policy optimization. RePO integrates an \emph{on-policy} update with an \emph{off-policy} replay-buffer update, increasing flexibility in the optimization process. An overview of RePO is shown in Figure~\ref{fig:method}. The objective function is defined as:
\begin{align*}
\mathcal{J}_{\mathrm{RePO}}(\theta;S)
&= \underbrace{\mathcal{J}_{\mathrm{on-policy}}(\theta)}_{\text{current samples}}
  + \underbrace{\mathcal{J}_{\mathrm{off-policy}}(\theta;S)}_{\text{replay samples}}.
\end{align*}
The replay strategy $S$ will be further detailed in Section~\textsection\ref{sec:off_policy_update}.
 
\subsection{On-Policy Update}
The on-policy part follows GRPO without applying any KL penalty, as suggested by \citep{yu2025dapo, yan2025learning, hu2025open}:
\begin{align*}
&\mathcal{J}_{\mathrm{on-policy}}(\theta)
= \mathbb{E}_{q\sim P(Q),\,\{o_i^{\mathrm{on}}\}_{i=1}^{G^{\mathrm{on}}}\sim\pi_{\theta_{\mathrm{old}}}(O\mid q)} \\
&\begin{aligned}[t]
\;    \frac{1}{{G^{\mathrm{on}}}}\sum_{i=1}^{G^{\mathrm{on}}}\frac{1}{\lvert o_i^{\mathrm{on}}\rvert}
       \sum_{t=1}^{\lvert o_i^{\mathrm{on}}\rvert}
       \min\bigl[\,&r_{i,t}^{\mathrm{on}}\,A_{i,t}^{\mathrm{on}},\, \\ 
    \quad \mathrm{clip}\bigl(&r_{i,t}^{\mathrm{on}},\,1-\epsilon,\,1+\epsilon\bigr)\,A_{i,t}^{\mathrm{on}}\bigr],
\end{aligned}
\end{align*}
where $$r_{i,t}^{\mathrm{on}}
= \frac{\pi_{\theta}\bigl(o_{i,t}^{\mathrm{on}}\mid q,\,o_{i, <t}^{\mathrm{on}}\bigr)}
       {\pi_{\theta_{\mathrm{old}}}\bigl(o_{i,t}^{\mathrm{on}}\mid q,\,o_{i, <t}^{\mathrm{on}}\bigr)}.$$
After each on-policy update, the sampled outputs and their generation probabilities are stored in a replay buffer $\mathcal{B}$ for subsequent off-policy updates.

\subsection{Off-Policy Update}
\label{sec:off_policy_update}
The off-policy part follows a similar structure as the on-policy part, but the data are retrieved from the replay buffer $\mathcal{B}$ containing previously generated outputs $o_t^{\mathrm{off}}$ along with their data-generating probabilities $\pi_{\theta_{\mathrm{off}}}$:  
\begin{align*}
&\mathcal{J}_{\mathrm{off-policy}}(\theta;S)
= \mathbb{E}_{\footnotesize
\mathrlap{
\begin{array}[t]{@{}l@{}}
q \sim P(Q), \\
\hspace{-3.5em}\{o_i^{\mathrm{off}},\, \pi_{\theta_{\mathrm{off}}}(o_i^{\mathrm{off}}\mid q)\}_{i=1}^{G^{\mathrm{off}}} \sim \mathcal{B}(q, S)
\end{array}
}
} \\
&\begin{aligned}[t]
\;  \frac{1}{{G^{\mathrm{off}}}}\sum_{i=1}^{G^{\mathrm{off}}}\frac{1}{\lvert o_i^{\mathrm{off}}\rvert}
       \sum_{t=1}^{\lvert o_i^{\mathrm{off}}\rvert}
       \min\bigl[\,&r_{i,t}^{\mathrm{off}}\,A_{i,t}^{\mathrm{off}},\, \\ 
    \quad\mathrm{clip}\bigl(&r_{i,t}^{\mathrm{off}},\,1-\epsilon,\,1+\epsilon\bigr)\,A_{i,t}^{\mathrm{off}}\bigr],
\end{aligned}
\end{align*}
where $$r_{i,t}^{\mathrm{off}} = \frac{\pi_{\theta}\bigl(o_{i,t}^{\mathrm{off}}\mid q,\,o_{i, <t}^{\mathrm{off}}\bigr)}{\pi_{\theta_{\mathrm{off}}}\bigl(o_{i,t}^{\mathrm{off}}\mid q,\,o_{i,<t}^{\mathrm{off}}\bigr)}.$$
\(\{o_i^{\mathrm{off}}, \pi_{\theta_{\mathrm{off}}}\bigl(o_i^{\mathrm{off}}\mid q\bigr)\}_{i=1}^{G^{\mathrm{off}}}\sim \mathcal{B}(q, S)\) denotes retrieving a group of previous outputs and the associated generation probability $\pi_{\theta_{\mathrm{off}}}$ for prompt $q$ based on replay strategy $S$. The off-policy component optimizes the current policy using suitable previous samples, increasing data efficiency. Additionally, the $\mathrm{clip}$ operation prevents excessive divergence between the current policy and the behavior policy that generated the retrieved samples.

\paragraph{Replay Strategy.}
RePO provides flexibility by allowing diverse retrieval strategies tailored to specific tasks and models. This part introduces several replay strategies as references.

\noindent
\underline{\textit{Full-scope.}}
This strategy retrieves all previous samples for the current optimization, increasing data usage. However, excessive reliance on past samples may interfere with the current policy, potentially complicating the optimization process.

\noindent
\underline{\textit{Recency-based.}}
To mitigate the above issue, this strategy retrieves the most recent $K$ samples, which aligns more closely with the current policy.

\noindent
\underline{\textit{Reward-oriented.}}
This strategy selects samples with the highest rewards, focusing on reinforcing desirable prior behaviors.

\noindent 
\underline{\textit{Variance-driven.}}
Inspired by the vanishing gradient issue in RL for LLMs \citep{razin2024vanishing, razin2025makes, xu2025not}, this strategy targets scenarios where low reward discriminability leads to weak gradient updates, despite the model being far from optimal. To address this, it retrieves a group of previous samples with the highest reward variance, providing a more substantial optimization signal.

\begin{algorithm*}[t]
  \caption{Replay-Enhanced Policy Optimization (RePO)}
  \textbf{Input:} Initial policy model $\pi_{\theta_{\text{init}}}$, reward model $R$, replay buffer $\mathcal{B}$, replay strategy $S$, task prompts $\mathcal{D}$, clipping parameter $\epsilon$, number of iterations $\mu$, number of epochs $N$, off-policy start epoch $E_{\mathrm{off}}$
  \begin{algorithmic}[1]
    \State Initialize policy model $\pi_\theta \gets \pi_{\theta_{\text{init}}}$
    \For{$\text{epoch} = 1, \ldots, N$}
      \For{$\text{step} = 1, \ldots, M$}
        \State Sample a batch $\mathcal{D}_b$ from $\mathcal{D}$
        \State Update the old policy model: $\pi_{\theta_{\mathrm{old}}} \gets \pi_{\theta}$

        \State Sample on-policy outputs $\{o_i^{\mathrm{on}}\}_{i=1}^{G^{\mathrm{on}}} \sim \pi_{\theta_{\mathrm{old}}} (\cdot \mid q)$ for each $q \in \mathcal{D}_b$
        \State Compute rewards $\{R(o_i^{\mathrm{on}})\}_{i=1}^{G^{\mathrm{on}}}$ using $R$
        \State Compute advantages $A_{i,t}^{\mathrm{on}}$ for $t$-th token of $o_i^{\mathrm{on}}$

        \If{$\text{epoch} \geq E_{\mathrm{off}}$}
            \State \textcolor{blue}{Sample off-policy outputs $\{o_i^{\mathrm{off}}, \pi_{\theta_{\mathrm{off}}}(o_i^{\mathrm{off}} \mid q)\}_{i=1}^{G^{\mathrm{off}}} \sim \mathcal{B}(q, S)$ for each $q \in \mathcal{D}_b$}
            \State \textcolor{blue}{Compute rewards $\{R(o_i^{\mathrm{off}})\}_{i=1}^{G^{\mathrm{off}}}$ using $R$}
            \State \textcolor{blue}{Compute advantages $A_{i,t}^{\mathrm{off}}$ for $t$-th token of $o_i^{\mathrm{off}}$}
          \For{$\text{iteration} = 1, \ldots, \mu$}
            \State \textcolor{blue}{Update policy model $\pi_{\theta}$ using $\mathcal{J}_{\mathrm{RePO}}(\theta; S)$}
          \EndFor
        \Else
          \For{$\text{iteration} = 1, \ldots, \mu$}
            \State Update policy model $\pi_{\theta}$ using $\mathcal{J}_{\mathrm{GRPO}}(\theta)$
          \EndFor
        \EndIf

        \State \textcolor{blue}{Update replay buffer $\mathcal{B}$ with $\{(q, o_i^{\mathrm{on}}, \pi_{\theta_{\mathrm{old}}}(o_i^{\mathrm{on}} \mid q))\}_{i=1}^{G^{\mathrm{on}}}$}
      \EndFor
    \EndFor
  \end{algorithmic}
  \textbf{Output:} Optimized policy model $\pi_\theta$
  \label{alg:repo-epoch-offpolicy}
\end{algorithm*}

\paragraph{What does the off-policy update do?}


If we calculate the gradient to the unclipped loss:
\begin{align*}
    &\nabla_{\theta}\mathcal{J}_{\mathrm{on-policy}}(\theta)_{i,t} \propto \nabla_{\theta} \left( r_{i,t}^{\mathrm{on}} \cdot A_{i,t}^{\mathrm{on}}  \right) \\
    &= r_{i,t}^{\mathrm{on}} \cdot A_{i,t}^{\mathrm{on}} \cdot \nabla_{\theta} \log \pi_{\theta}\bigl(o_{i,t}^{\mathrm{on}} \mid q,\, o_{i, <t}^{\mathrm{on}}\bigr).
\end{align*}
\begin{align*}
    &\nabla_{\theta}\mathcal{J}_{\mathrm{off-policy}}(\theta;S)_{i,t} \propto \nabla_{\theta} \left( r_{i,t}^{\mathrm{off}} \cdot A_{i,t}^{\mathrm{off}}  \right) \\
    &= r_{i,t}^{\mathrm{off}} \cdot A_{i,t}^{\mathrm{off}} \cdot \nabla_{\theta} \log \pi_{\theta}\bigl(o_{i,t}^{\mathrm{off}} \mid q,\, o_{i, <t}^{\mathrm{off}}\bigr).
\end{align*}
Assuming the two losses take the same data, i.e., $A_{i,t}^{\text{on}} = A_{i,t}^{\text{off}}$ and $\log \pi_{\theta}(o_{i,t}^{\text{on}} \mid q, o_{i,<t}^{\text{on}}) = \log \pi_{\theta}(o_{i,t}^{\text{off}} \mid q, o_{i,<t}^{\text{off}})$, and we only perform one gradient step for each group (common in practice), i.e.,  $\theta = \theta_{\text{old}} \rightarrow r_{i,t}^{\text{on}} = 1$. Thus,
\begin{equation*}
    \frac{\nabla_{\theta} \mathcal{J}_{\text{off}}(\theta; S)_{i,t}}{\nabla_{\theta} \mathcal{J}_{\text{on}}(\theta)_{i,t}} = r_{i,t}^{\text{off}} = \frac{\pi_{\theta}\bigl(o_{i,t}^{\mathrm{off}}\mid q,\,o_{i, <t}^{\mathrm{off}}\bigr)}{\pi_{\theta_{\mathrm{off}}}\bigl(o_{i,t}^{\mathrm{off}}\mid q,\,o_{i,<t}^{\mathrm{off}}\bigr)}.
\end{equation*}
In other words, compared to the standard on-policy loss $\mathcal{J}_{\text{on}}$, the off-policy loss $\mathcal{J}_{\text{off}}$ can be interpreted as:  
(1) using off-policy data,  
(2) applying the standard on-policy GRPO loss, and  
(3) scaling it by $r_{i,t}^{\mathrm{off}}$: the loss is downweighted when the current policy $\pi_\theta$ assigns low probability to the data compared to the behavior policy $\pi_{\theta_\mathrm{off}}$, and upweighted otherwise.  
\textbf{As a result, although off-policy loss reuses past samples, samples unlikely under the current policy contribute little to learning, preventing them from reversing the policy’s progress.}

\subsection{Advantage Estimation}
Given the rewards
\begin{align*}
\mathcal{G}^{\mathrm{on}}
= \{R(o_i^{\mathrm{on}})\}_{i=1}^{G^{\mathrm{on}}},
\qquad
\mathcal{G}^{\mathrm{off}}
= \{R(o_i^{\mathrm{off}})\}_{i=1}^{G^{\mathrm{off}}},
\end{align*}
\ours\ estimates advantages separately as follows:
\begin{align*}
A_i^{\mathrm{on}}
&= \frac{R(o_i^{\mathrm{on}})
   - \mathrm{mean}(\mathcal{G}^{\mathrm{on}})}
   {\mathrm{std}(\mathcal{G}^{\mathrm{on}})},\\
A_i^{\mathrm{off}}
&= \frac{R(o_i^{\mathrm{off}})
   - \mathrm{mean}(\mathcal{G}^{\mathrm{off}})}
   {\mathrm{std}(\mathcal{G}^{\mathrm{off}})}.
\end{align*}
This strategy promotes separate updates for on-policy and off-policy experiences, reducing potential interference. We also explore a mixed strategy that integrates both sample types, detailed in \textsection\ref{sec:analysis}.

\section{Experiments}
\begin{table*}[ht]
  \centering
  \renewcommand{\arraystretch}{1.0}
  \resizebox{\textwidth}{!}{%
    \begin{tabular}{lcccccccc}
      \toprule
      \textbf{Training Strategy} & \textbf{GSM8K} & \textbf{MATH-500} & \textbf{Olympiad} &
      \textbf{Minerva} & \textbf{AIME24} & \textbf{AIME25} & \textbf{AMC23} & \textbf{Avg.} \\
      \midrule
      \multicolumn{9}{c}{\textit{Qwen2.5-Math-1.5B}} \\
      \midrule
      GRPO   &  $9.8$ & $29.6$ & $52.2$ & $3.7$  & $2.9$  & $\mathbf{5.2}$ & $18.2$ & $17.4$ \\
      \ours  &  $\mathbf{66.5}$ & $\mathbf{52.0}$ & $\mathbf{64.4}$ & $\mathbf{11.4}$ & $\mathbf{7.7}$ & $\mathbf{5.2}$ & $\mathbf{43.7}$ & $\mathbf{35.8}$ \\
      \midrule
      \multicolumn{9}{c}{\textit{Qwen2.5-Math-1.5B-Instruct}} \\
      \midrule
      GRPO   &  $\mathbf{86.5}$ & $59.4$ & $\mathbf{82.2}$ & $\mathbf{17.3}$ & $\mathbf{10.7}$ & $9.3$  & $55.5$ & $45.8$ \\
      \ours  &  $86.1$ & $\mathbf{61.6}$ & $82.0$ & $16.9$ & $10.2$ & $\mathbf{9.7}$ & $\mathbf{56.5}$ & $\mathbf{46.1}$ \\
      \midrule
      \multicolumn{9}{c}{\textit{Qwen2.5-Math-7B}} \\
      \midrule
      GRPO   &  $\mathbf{88.5}$ & $\mathbf{63.2}$ & $76.8$ & $\mathbf{19.9}$ & $17.5$ & $8.3$  & $54.8$ & $47.0$ \\
      \ours  &  $88.2$ & $62.2$ & $\mathbf{81.9}$ & $18.4$ & $\mathbf{21.8}$ & $\mathbf{10.5}$ & $\mathbf{59.9}$ & $\mathbf{49.0}$ \\
      \midrule
      \multicolumn{9}{c}{\textit{Qwen2.5-Math-7B-Instruct}} \\
      \midrule
      GRPO   &  $90.4$ & $53.0$ & $49.1$ & $16.5$ & $2.6$  & $0.8$  & $38.0$ & $35.8$ \\
      \ours  &  $\mathbf{93.7}$ & $\mathbf{62.2}$ & $\mathbf{64.4}$ & $\mathbf{19.5}$ & $\mathbf{5.5}$ & $\mathbf{3.2}$ & $\mathbf{46.3}$ & $\mathbf{42.1}$ \\
      \midrule
      \multicolumn{9}{c}{\textit{Qwen3-1.7B}} \\
      \midrule
      GRPO   &  $87.1$ & $54.2$ & $55.0$ & $18.8$ & $5.8$  & $12.3$ & $43.0$ & $39.5$ \\
      \ours  &  $\mathbf{88.2}$ & $\mathbf{59.6}$ & $\mathbf{63.1}$ & $\mathbf{20.2}$ & $\mathbf{12.3}$ & $\mathbf{13.8}$ & $\mathbf{47.7}$ & $\mathbf{43.6}$ \\
      \bottomrule
    \end{tabular}%
  }
  \caption{\textbf{Comparison of GRPO and \ours\ on math reasoning benchmarks.} Highlighted entries indicate the best performance per model. \ours\ consistently outperforms GRPO on average across evaluated models.}
  \label{tab:training-strategy-comparison}
\end{table*}
\begin{table*}[ht]
  \centering
  \renewcommand{\arraystretch}{1.0}
  \resizebox{\textwidth}{!}{%
    \begin{tabular}{lcccccccc}
      \toprule
      \textbf{Method} & \textbf{GSM8K} & \textbf{MATH-500} & \textbf{Olympiad} &
      \textbf{Minerva} & \textbf{AIME24} & \textbf{AIME25} & \textbf{AMC23} & \textbf{Avg.} \\
      \midrule
      \multicolumn{9}{c}{\textit{Qwen2.5-Math-1.5B}} \\
      \midrule
      Dr.\ GRPO & $\mathbf{65.6}$ & $\mathbf{50.0}$ & $33.9$ & $10.7$ & $6.3$ & $\mathbf{5.6}$ & $44.1$ & $30.9$ \\
      RePO (Dr.\ GRPO) & $62.0$ & $47.6$ & $\mathbf{47.5}$ & $\mathbf{11.0}$ & $\mathbf{8.3}$ & $4.5$ & $\mathbf{44.7}$ & $\mathbf{32.2}$ \\
      \midrule
      \multicolumn{9}{c}{\textit{Qwen3-1.7B}} \\
      \midrule
      Dr.\ GRPO & $85.8$ & $55.6$ & $51.3$ & $\mathbf{21.3}$ & $5.7$ & $\mathbf{9.0}$ & $43.2$ & $38.8$ \\
      RePO (Dr.\ GRPO) & $\mathbf{87.5}$ & $\mathbf{59.2}$ & $\mathbf{66.3}$ & $20.6$ & $\mathbf{11.0}$ & $8.8$ & $\mathbf{47.7}$ & $\mathbf{43.0}$ \\
      \bottomrule
    \end{tabular}%
  }
  \caption{\textbf{Comparison of Dr.\ GRPO and RePO (Dr.\ GRPO) on math reasoning benchmarks}, with RePO (Dr.\ GRPO) indicating the RePO variant built upon Dr.\ GRPO, evaluated using Qwen2.5-Math-1.5B and Qwen3-1.7B.}
  \label{tab:rl-method-comparison}
\end{table*}
\begin{table*}
  \centering
  \renewcommand{\arraystretch}{1.0}
  \resizebox{\textwidth}{!}{%
    \begin{tabular}{lccccccc}
      \toprule
      \textbf{Training Strategy} & \textbf{MMLU-Pro} & \textbf{ARC-Easy} & \textbf{ARC-Challenge} &
      \textbf{GPQA-Diamond} & \textbf{BBH} & \textbf{IFEval} & \textbf{Avg.} \\
      \midrule
      \multicolumn{8}{c}{\textit{Qwen2.5-Math-1.5B}} \\
      \midrule
      GRPO  & $11.4$ & $38.3$ & $\mathbf{31.7}$ & $26.0$ & $34.8$ & $\mathbf{20.3}$ & $27.1$ \\
      \ours & $\mathbf{12.6}$ & $\mathbf{38.5}$ & $30.5$ & $\mathbf{26.3}$ & $\mathbf{36.9}$ & $\mathbf{20.3}$ & $\mathbf{27.5}$ \\
      \midrule
      \multicolumn{8}{c}{\textit{Qwen2.5-Math-1.5B-Instruct}} \\
      \midrule
      GRPO  & $18.0$ & $\mathbf{40.0}$ & $\mathbf{31.9}$ & $\mathbf{27.3}$ & $\mathbf{37.3}$ & $23.6$ & $29.5$ \\
      \ours & $\mathbf{18.1}$ & $\mathbf{40.0}$ & $31.0$ & $\mathbf{27.3}$ & $37.1$ & $\mathbf{24.3}$ & $\mathbf{29.6}$ \\
      \midrule
      \multicolumn{8}{c}{\textit{Qwen2.5-Math-7B}} \\
      \midrule
      GRPO  & $27.4$ & $47.5$ & $38.4$ & $20.7$ & $43.6$ & $31.5$ & $34.9$ \\
      \ours & $\mathbf{30.3}$ & $\mathbf{50.3}$ & $\mathbf{38.6}$ & $\mathbf{28.3}$ & $\mathbf{44.5}$ & $\mathbf{31.9}$ & $\mathbf{37.3}$ \\
      \midrule
      \multicolumn{8}{c}{\textit{Qwen2.5-Math-7B-Instruct}} \\
      \midrule
      GRPO  & $17.0$ & $47.3$ & $\mathbf{38.0}$ & $\mathbf{29.3}$ & $34.6$ & $32.1$ & $33.1$ \\
      \ours & $\mathbf{23.1}$ & $\mathbf{48.4}$ & $37.7$ & $27.8$ & $\mathbf{38.1}$ & $\mathbf{32.4}$ & $\mathbf{34.6}$ \\
      \midrule
      \multicolumn{8}{c}{\textit{Qwen3-1.7B}} \\
      \midrule
      GRPO  & $\mathbf{11.7}$ & $38.3$ & $\mathbf{34.8}$ & $\mathbf{26.3}$ & $\mathbf{28.5}$ & $47.4$ & $31.2$ \\
      \ours & $\mathbf{11.7}$ & $\mathbf{38.5}$ & $34.7$ & $\mathbf{26.3}$ & $28.4$ & $\mathbf{48.8}$ & $\mathbf{31.4}$ \\
      \bottomrule
    \end{tabular}%
  }
  \caption{\textbf{Comparison of GRPO and \ours\ on general reasoning benchmarks.} Highlighted entries indicate the best performance per model. RePO consistently outperforms GRPO on average across evaluated models.}
  \label{tab:training-strategy-general}
\end{table*}

\begin{figure*}[h]
  \centering
  \includegraphics[width=1.0\textwidth]{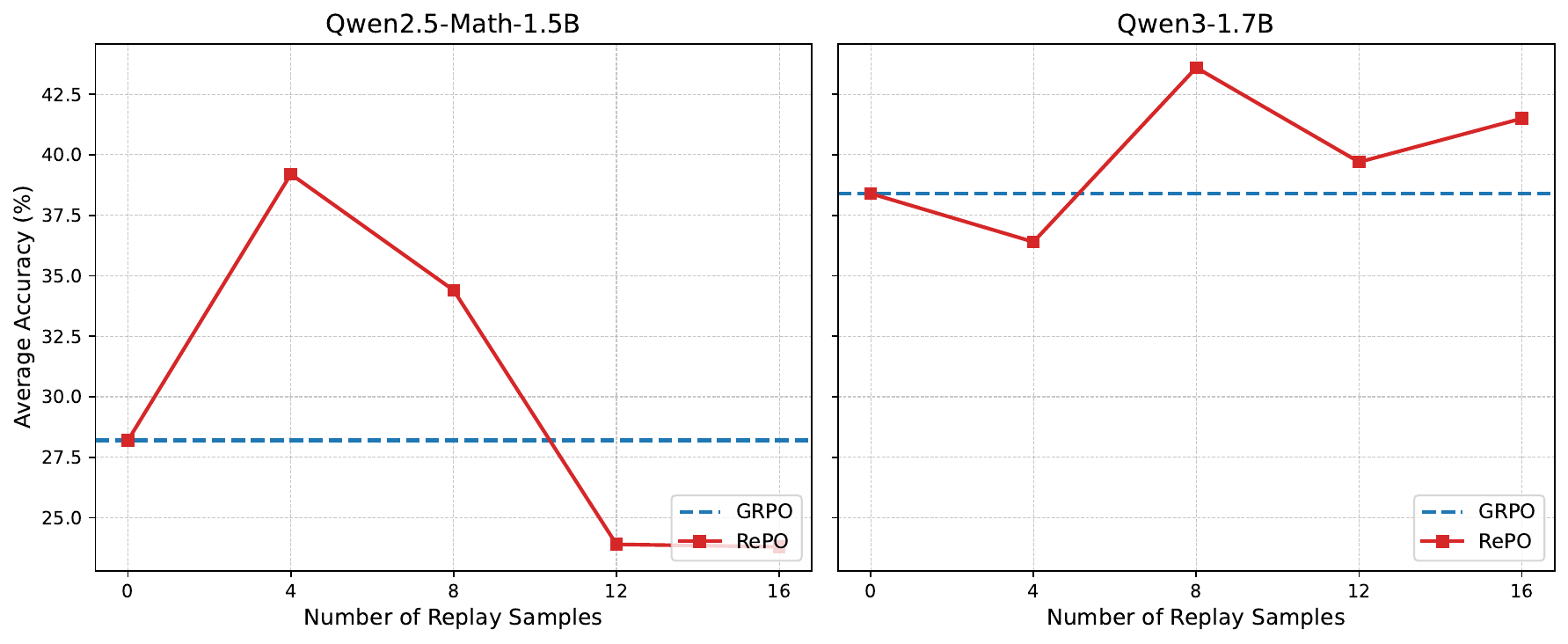}
  \caption{\textbf{Comparison of average accuracy between GRPO and RePO under varying numbers of off-policy (Replay) samples} across seven math reasoning benchmarks: GSM8K, MATH-500, Olympiad, Minerva, AIME24, AIME25, and AMC23, using Qwen2.5-Math-1.5B (left) and Qwen3-1.7B (right) with 8 on-policy samples.}
  \label{fig:effect-of-replay-numbers}
\end{figure*}

\begin{table*}[ht]
  \centering
  \renewcommand{\arraystretch}{1.0}
  \resizebox{\textwidth}{!}{%
    \begin{tabular}{lcccccccc}
      \toprule
      \textbf{Advantage Estimation} & \textbf{GSM8K} & \textbf{MATH-500} & \textbf{Olympiad} &
      \textbf{Minerva} & \textbf{AIME24} & \textbf{AIME25} & \textbf{AMC23} & \textbf{Avg.} \\
      \midrule
      \multicolumn{9}{c}{\textsc{On-policy:}~4 \quad \textsc{Off-policy:}~4} \\
      \midrule
      \textsc{Mixed} & $63.9$ & $23.6$ & $14.2$ & $10.3$ & $0.0$ & $0.0$ & $15.1$ & $18.2$ \\
      \textsc{Split} & $\mathbf{77.2}$ & $\mathbf{39.4}$ & $\mathbf{25.7}$ & $\mathbf{11.8}$ & $\mathbf{0.1}$ & $\mathbf{2.2}$ & $\mathbf{24.6}$ & $\mathbf{25.9}$ \\
      \midrule
      \multicolumn{9}{c}{\textsc{On-policy:}~8 \quad \textsc{Off-policy:}~8} \\
      \midrule
      \textsc{Mixed} & $79.7$ & $42.4$ & $28.2$ & $11.8$ & $0.1$ & $2.5$ & $25.8$ & $27.2$ \\
      \textsc{Split} & $\mathbf{88.2}$ & $\mathbf{59.6}$ & $\mathbf{63.1}$ & $\mathbf{20.2}$ & $\mathbf{12.3}$ & $\mathbf{13.8}$ & $\mathbf{47.7}$ & $\mathbf{43.6}$ \\
      \bottomrule
    \end{tabular}%
  }
  \caption{\textbf{Comparison of split and mixed advantage estimation strategies} using Qwen3-1.7B on mathematical reasoning benchmarks, with both on-policy and off-policy sample numbers set to $4$ or $8$.}
  \label{tab:advantage-computation}
\end{table*}
\begin{table*}[h!]
  \centering
  \renewcommand{\arraystretch}{1.0}
  \resizebox{\textwidth}{!}{%
    \begin{tabular}{lcccccccc}
      \toprule
      \textbf{Replay Strategy} & \textbf{GSM8K} & \textbf{MATH-500} & \textbf{Olympiad} &
      \textbf{Minerva} & \textbf{AIME24} & \textbf{AIME25} & \textbf{AMC} & \textbf{Avg.} \\
      \midrule
      \multicolumn{9}{c}{\textit{Qwen2.5-Math-1.5B}} \\
      \midrule
      No             & $40.3$ & $45.8$ & $57.4$ & $7.0$  & $5.4$  & $5.5$  & $36.3$ & $28.2$ \\
      Random         & $51.7$ & $47.6$ & $53.5$ & $9.6$  & $6.6$  & $5.5$  & $39.8$ & $30.6$ \\
      Full-scope            & $25.4$ & $35.4$ & $62.5$ & $8.1$  & $4.8$  & $4.7$  & $25.5$ & $23.8$ \\
      Recency-based    & $\mathbf{66.5}$ & $\mathbf{52.0}$ & $\mathbf{64.4}$ & $\mathbf{11.4}$ & $\mathbf{7.7}$ & $5.2$           & $\mathbf{43.7}$ & $\mathbf{35.8}$ \\
      Reward-oriented & $41.5$ & $45.8$ & $56.3$ & $11.0$ & $5.1$  & $4.4$  & $34.5$ & $28.4$ \\
      Variance-driven   & $44.5$ & $45.6$ & $60.7$ & $8.1$  & $5.6$  & $\mathbf{6.8}$ & $34.4$ & $29.4$ \\
      \midrule
      \multicolumn{9}{c}{\textit{Qwen3-1.7B}} \\
      \midrule
      No             & $84.9$ & $53.6$ & $55.9$ & $17.6$ & $4.3$  & $9.2$  & $43.6$ & $38.4$ \\
      Random         & $85.6$ & $51.2$ & $46.8$ & $16.5$ & $2.9$  & $6.4$  & $38.2$ & $35.4$ \\
      Full-scope            & $87.6$ & $\mathbf{59.6}$ & $59.5$ & $18.0$ & $8.0$  & $11.5$ & $46.4$ & $41.5$ \\
      Recency-based    & $86.1$ & $58.0$ & $63.0$ & $\mathbf{21.3}$ & $10.7$ & $12.9$ & $43.9$ & $42.3$ \\
      Reward-oriented & $\mathbf{88.2}$ & $\mathbf{59.6}$ & $\mathbf{63.1}$ & $20.2$ & $\mathbf{12.3}$ & $\mathbf{13.8}$ & $\mathbf{47.8}$ & $\mathbf{43.6}$ \\
      Variance-driven   & $85.8$ & $54.6$ & $46.9$ & $17.3$ & $3.2$  & $9.1$  & $38.3$ & $36.5$ \\
      \bottomrule
    \end{tabular}%
  }
  \caption{\textbf{Impact of replay strategies on performance across math reasoning benchmarks}. All experiments use $8$ on-policy and $8$ off-policy samples. Highlighted entries denote the best performance for each model.}
  \label{tab:replay-strategy}
\end{table*}

\subsection{Evaluation Setup}
We primarily compare \ours\ with GRPO, a leading RL method for optimizing reasoning capabilities in LLMs. To evaluate performance, we focus on seven widely used math reasoning benchmarks: GSM8K \citep{cobbe2021training}, MATH-500 \citep{hendrycks2021measuring}, OlympiadBench \citep{he2024olympiadbench}, Minerva \citep{lewkowycz2022solving}, AIME24, AIME25, and AMC \citep{li2024numinamath}. We report avg@32 for AIME24, AIME25, and AMC due to their small test set sizes, and pass@1 for the remaining benchmarks. The decoding hyperparameters are set to a temperature of $0.2$ and a $\text{top-p}$ of $0.95$ \citep{Holtzman2020The}. Additionally, we evaluate the general reasoning capabilities of \ours\ using MMLU-Pro \citep{wang2024mmlupro}, ARC \citep{clark2018think}, GPQA \citep{rein2024gpqa}, BBH \citep{suzgun2022challenging}, and IFEval \citep{zhou2023instruction}.

\subsection{Implementation}
\paragraph{Dataset.} We concentrate on mathematical reasoning tasks and adopt the DeepMath dataset \citep{he2025deepmath} for training, which comprises a diverse and challenging set of problems. With an emphasis on the RL setting, we use only the problems and their corresponding ground-truth answers. Due to computational constraints, we randomly sample a subset of $1024$ examples for training.

\paragraph{Model.} We conduct experiments using the Qwen series LLMs \citep{yang2024qwen2, yang2025qwen3technicalreport}, which have shown strong performance in complex reasoning tasks \citep{zeng2025simplerl, qwq32b}. To assess the general applicability of \ours, we include multiple models, such as Qwen2.5-Math-1.5B, Qwen2.5-Math-1.5B-Instruct, Qwen2.5-Math-7B, Qwen2.5-Math-7B-Instruct, and Qwen3-1.7B. 

\paragraph{Training.} A comprehensive description of the training process is provided in Appendix~\ref{sec:app_training_details}.

\subsection{Main Results}

\paragraph{Mathematical Reasoning.} 
Table~\ref{tab:training-strategy-comparison} compares the performance of GRPO and \ours\ on seven widely used mathematical reasoning benchmarks using five LLMs. The optimal replay strategy is model-specific and task-dependent. To prevent overfitting and demonstrate robustness, we universally adopt Recency-based replay for all base models (Qwen2.5-Math-1.5B, Qwen2.5-Math-7B) and Reward-oriented replay for all instruct models (Qwen2.5-Math-1.5B-Instruct, Qwen2.5-Math-7B-Instruct, Qwen3-1.7B). This configuration is uniformly maintained across all experiments involving \ours in this paper. A detailed comparison of replay strategies is provided in Section~\textsection\ref{sec:analysis}, Table~\ref{tab:replay-strategy}. We conjecture that the primary advantage of \ours\ lies in its ability to leverage outputs from previous steps: by performing policy optimization over both on-policy and off-policy samples for each prompt, \ours\ increases the diversity of training data and effectively mitigates overfitting.

\paragraph{General Reasoning.}
Table~\ref{tab:training-strategy-general} presents the comparison of GRPO and \ours\ across six general reasoning tasks using five LLMs. \ours\ achieves slight yet consistent improvements over GRPO, with absolute gains in average performance ranging from $0.2$ to $2.4$ points, demonstrating stronger generalization across diverse reasoning tasks.

\subsection{Analysis}
\label{sec:analysis}

\paragraph{Generality across RL Methods.}
In this paper, \ours\ is primarily implemented based on GRPO; however, the core concept of replaying can be applied to other RL methods as well. To verify this, we conduct experiments using Dr. GRPO~\citep{liu2025understanding}, a GRPO variant designed to reduce length and difficulty biases. As shown in Table~\ref{tab:rl-method-comparison}, \ours\ applied to Dr. GRPO also outperforms the original Dr. GRPO, achieving absolute average performance gains of $1.3$ and $4.2$ points for Qwen2.5-Math-1.5B and Qwen3-1.7B, respectively, across seven mathematical reasoning benchmarks.

\paragraph{Advantage Estimation.}
\ours\ employs a Split strategy that estimates advantages separately for on-policy and off-policy samples (\textsection\ref{sec:repo}), while an alternative is to mix both sample types as a unified group for group relative advantage estimation (\textsection\ref{sec:grpo}). While the Mixed approach provides greater flexibility by allowing off-policy samples to influence on-policy updates, it may also introduce interference between the two. As shown in Table~\ref{tab:advantage-computation}, the Split strategy consistently outperforms the Mixed strategy across all mathematical reasoning benchmarks, achieving average performance gains of $7.7$ and $16.4$ points under different sample settings on Qwen3-1.7B. These results underscore the importance of separating on-policy and off-policy samples during advantage estimation to reduce mutual interference. Accordingly, we employ the Split strategy throughout this work and leave exploration of the Mixed approach for future research.

\paragraph{Impact of Replay Strategy.}
We evaluate the impact of different replay strategies using Qwen2.5-Math-1.5B and Qwen3-1.7B, comparing the four proposed strategies (\textsection\ref{sec:off_policy_update}) with two baseline approaches: no replay and random replay. As shown in Table~\ref{tab:replay-strategy}, Recency-based and Reward-oriented strategies consistently achieve superior performance. The former retrieves samples that align more closely with the current policy, while the latter prioritizes high-reward samples, thereby reinforcing desirable behaviors.

\paragraph{Impact of Replay Numbers.}
We examine how the number of off-policy samples influences performance. As shown in Figure~\ref{fig:effect-of-replay-numbers}, performance initially improves with more off-policy samples but eventually declines. This pattern is expected, as fewer samples may provide insufficient training signals, while excessive samples can introduce noise due to the finite number of off-policy samples.

\paragraph{Computational Cost.}
\begin{table}[htp]
  \centering
  \renewcommand{\arraystretch}{1.0}
  \resizebox{1.0\linewidth}{!}{
  \begin{tabular}{lcc}
    \toprule
    \textbf{Method} & \textbf{Avg. Acc.} & \textbf{Rel. Time} \\
    \midrule
    \multicolumn{3}{c}{\textsc{On-policy:}~4} \\
    \midrule
    GRPO                       & $21.1$ & $\times\mathbf{1.00}$ \\
    \ours\ (\textsc{Off-policy:}~4) & $\mathbf{25.9}$ & $\times 1.15$ \\
    \midrule
    \multicolumn{3}{c}{\textsc{On-policy:}~8} \\
    \midrule
    GRPO                       & $39.5$ & $\times\mathbf{2.02}$ \\
    \ours\ (\textsc{Off-policy:}~8) & $\mathbf{43.6}$ & $\times 2.30$ \\
    \bottomrule
  \end{tabular}
  }
  \caption{\textbf{Average math reasoning accuracy and relative training time for GRPO and RePO} on Qwen3-1.7B. \ours\ uses 4 and 8 off-policy samples; training time is normalized to GRPO with 4 on-policy samples.}
  \label{tab:avg-computaional-cost}
\end{table}


We assess the computational cost of \ours\ against GRPO on Qwen3-1.7B. As shown in Table \ref{tab:avg-computaional-cost}, \ours\ incurs a $15\%$ relative increase in computational cost across various sample settings. Despite this, it achieves absolute gains of $4.8$ and $4.1$ points in average performance over GRPO across seven math reasoning benchmarks.

\paragraph{Rationale for Effectiveness.}

\begin{table}[ht]
  \centering
  \resizebox{1.0\linewidth}{!}{
  \begin{tabular}{lcc}
    \toprule
    \textbf{Method} & \textbf{Avg. Acc.} & \textbf{Eff. Steps} \\
    \midrule
    \multicolumn{3}{c}{\textsc{On-policy:}~4} \\
    \midrule
    GRPO                       & $21.1$            & $13.0\%$          \\
    \ours\ (\textsc{Off-policy:}~4) & $\mathbf{25.9}$  & $\mathbf{14.3\%}$ ($+10.0\%$) \\
    \midrule
    \multicolumn{3}{c}{\textsc{On-policy:}~8} \\
    \midrule
    GRPO                       & $39.5$            & $31.2\%$          \\
    \ours\ (\textsc{Off-policy:}~8) & $\mathbf{43.6}$  & $\mathbf{46.1\%}$ ($+47.8\%$) \\
    \bottomrule
  \end{tabular}
  }
  \caption{\textbf{Average math reasoning accuracy and effective-step percentage for GRPO versus RePO} on Qwen3-1.7B. Relative improvements in effective-step percentage are shown in parentheses.}
  \label{tab:effective-steps}
\end{table}

We conjecture that the effectiveness of \ours\ primarily arises from optimizing the policy using a larger set of outputs per prompt, which helps reduce overfitting. Additionally, \ours\ increases the number of effective optimization steps. In GRPO (\textsection\ref{sec:grpo}), when all on-policy samples in a step receive the same reward, either all $1$ or all $0$, both the advantage and gradient become zero, resulting in an ineffective optimization step.\footnote{Here, we assume that each step consists of a single prompt, with optimization based solely on its sampled outputs. According to the advantage estimation method and objective function in GRPO (\textsection\ref{sec:grpo}), both the advantage and gradient would be zero in this case.} \ours\ mitigates this limitation by applying suitable replay strategies to retrieve previous outputs and incorporating more samples per step. As shown in Table \ref{tab:effective-steps}, \ours\ achieves a $10.0\%$ and $47.8\%$ relative increase in the number of effective steps when the number of on-policy and off-policy samples are both set to $4$ and $8$, respectively, compared with GRPO, demonstrating its effectiveness.
\section{Related Work}
Reinforcement learning (RL) has emerged as a critical step in optimizing LLMs. Early studies employing PPO \citep{schulman2017proximal} focus on tasks such as summarization \citep{stiennon2020learning}, aligning LLMs with human values \citep{bai2022training, ouyang2022training}, and reasoning \citep{havrilla2024teaching}. However, PPO relies on a large value model for advantage estimation, leading to increased memory and computational costs. To mitigate value model dependency and provide alternative advantage estimation, ReMax \citep{li2024remax} applies greedy search to generate a single output for each prompt and sets its reward as the baseline. RLOO \citep{ahmadian2024back} samples multiple outputs per prompt, and uses the mean reward of other outputs as the baseline. REINFORCE++ \citep{hu2025reinforceefficientrlhfalgorithm} uses the mean reward of a global batch as the baseline. 

GRPO \citep{shao2024deepseekmath} samples multiple outputs per prompt and estimate advantage by normalizing their rewards, effectively reducing memory usage while improving the reasoning capabilities of LLMs. Recent studies have identified limitations in GRPO, including length and difficulty biases \citep{liu2025understanding} and entropy collapse \citep{yu2025dapo}, and proposed corresponding mitigation strategies. LUFFY \citep{yan2025learning} extends GRPO by incorporating off-policy samples generated by more advanced models, such as DeepSeek-R1 \citep{guo2025deepseek}. In contrast, \ours\ enhances GRPO by using off-policy samples generated from previous iterations, eliminating the reliance on more advanced models and improving data efficiency. Additionally, we introduce several simple yet effective replay strategies to select suitable samples, further increasing the flexibility of policy optimization.
\section{Conclusion}
This study proposes Replay-Enhanced Policy Optimization (RePO), which retrieves previously sampled off-policy outputs for current policy optimization. This approach optimizes the policy using a broader set of outputs per prompt, reducing overfitting and increasing flexibility. Experiments across five LLMs, seven mathematical reasoning benchmarks, and six general reasoning tasks validate the effectiveness of RePO. We expect this work to contribute to RL optimization for advancing LLMs.
\section*{Limitations}
This work has limitations that warrant further investigation.

\paragraph{Model.}
The experiments are limited to LLMs with up to $7B$ parameters, leaving larger models unexplored due to computational constraints. Assessing RePO on more capable models remains a valuable direction.

\paragraph{Training.}
Hyperparameters such as the learning rate, number of replay samples, clipping ratio, and a coefficient for adjusting the weight of the off-policy update were not thoroughly investigated due to computational cost. Further analysis of these factors is important.

\iftoggle{isReview}{}{
\section*{Author Contributions}
\textbf{Siheng Li} and \textbf{Zhanhui Zhou} made valuable contributions to the design of RePO. \textbf{Zhanhui Zhou} initiated the idea of integrating a replay buffer into GRPO in discussion with \textbf{Siheng Li} and contributed to the loss analysis. \textbf{Siheng Li} proposed the RePO loss, introduced key designs such as mixing on- and off-policy losses, led the experiments, and wrote most of the paper. \textbf{Other authors} supervised and managed the group.
}

\bibliography{custom}

\appendix

\section{Appendix}
\label{sec:appendix}

\subsection{Training Details}
\label{sec:app_training_details}
Both GRPO and \ours\ are trained for three epochs under identical configurations. In RePO, the off-policy update is applied only in the final epoch. The global batch size is $32$, with both on-policy and off-policy sample numbers set to $8$ per step. For GRPO, the number of training examples per step remains $32$ across all epochs, while in RePO, it increases to $64$ in the last epoch. The learning rate is $1e-6$, following a cosine decay schedule. The maximum token lengths for prompts and completions are configured as $512$ and $1024$, respectively. All experiments utilize $8$ NVIDIA A100 GPUs, with $4$ allocated for policy optimization and $4$ for output sampling using VLLM \citep{kwon2023efficient}. The reward function is based on Math-Verify\footnote{\url{https://github.com/huggingface/Math-Verify}}, assigning a reward of $1$ for outputs with correct final answers and $0$ otherwise.

\end{document}